\documentclass[11pt]{article}

\usepackage[preprint]{acl}

\usepackage{times}
\usepackage{latexsym}

\usepackage[T1]{fontenc}
\usepackage[utf8]{inputenc}

\usepackage{microtype}

\usepackage{inconsolata}

\usepackage{xcolor}         % colors

\usepackage{graphicx}
\usepackage{enumitem}
\usepackage{booktabs}
\usepackage{amsmath,amsfonts}
\usepackage{multirow}
\usepackage{multicol}
\usepackage{makecell}
\usepackage{fvextra}
\usepackage[table]{xcolor}
\usepackage[most]{tcolorbox}
\newtcolorbox{textbox}[1][]{%
    enhanced,
    colback=gray!5,      % Very light gray background
    colframe=gray!60,    % Darker gray border (not stark black)
    boxrule=0.5pt,       % Thin border
    sharp corners,       % ACL style usually prefers sharp corners
    left=1pt, right=1pt, top=1pt, bottom=1pt, % Padding
    fontupper=\small,    % Often examples look better slightly smaller
    breakable,           % Allows box to split across pages
    #1                   % Allows passing extra options on the fly
}

\DefineVerbatimEnvironment{promptblock}{Verbatim}{
  breaklines=true,
  breakanywhere=true,
  fontsize=\fontsize{8}{8},
  frame=single,
  framesep=1mm,
  breaksymbolleft={}
}

\title{TAPO: Translation Augmented Policy Optimization for Multilingual Mathematical Reasoning}

\author{
    \textbf{Xu Huang\textsuperscript{1}},
    \textbf{Zhejian Lai\textsuperscript{1}},
    \textbf{Zixian Huang\textsuperscript{2}},
    \textbf{Jiajun Chen\textsuperscript{1}},
    \textbf{Shujian Huang\textsuperscript{1}\thanks{Corresponding author}}
\\
    \textsuperscript{1}National Key Laboratory for Novel Software Technology, Nanjing University \\
    \textsuperscript{2}Shanghai Artificial Intelligence Laboratory
\\
    \texttt{\{xuhuang,laizj\}@smail.nju.edu.cn, huangzixian@pjlab.org.cn}\\
    \texttt{\{chenjj,huangsj\}@nju.edu.cn}
}

\begin{document}
\maketitle

\begin{abstract}
Large Language Models (LLMs) have demonstrated remarkable proficiency in English mathematical reasoning, yet a significant performance disparity persists in multilingual contexts, largely attributed to deficiencies in language understanding.
To bridge this gap, we introduce Translation-Augmented Policy Optimization (TAPO), a novel reinforcement learning framework built upon GRPO.
TAPO enforces an explicit alignment strategy where the model leverages English as a pivot and follows an understand-then-reason paradigm.
Crucially, we employ a step-level relative advantage mechanism that decouples understanding from reasoning, allowing the integration of translation quality rewards without introducing optimization conflicts.
Extensive experiments reveal that TAPO effectively synergizes language understanding with reasoning capabilities and is compatible with various models.
It outperforms baseline methods in both multilingual mathematical reasoning and translation tasks, while generalizing well to unseen languages and out-of-domain tasks.
\end{abstract}

\section{Introduction}

Large language models (LLMs)~\cite{hurst2024gpt, comanici2025gemini, DeepSeekAI2024DeepSeekV3TR, yang2025qwen3} have achieved great progress in English tasks like mathematical reasoning~\cite{jaech2024openai, guo2025deepseek}, even surpassing humans in competitions~\cite{liu2025deepseek}. However, a significant multilingual disparity still exists~\cite{huang-etal-2025-benchmax, huang-etal-2023-languages}.
Driven by the imbalance in the quantity and quality of training corpora across different languages, LLMs exhibit uneven language-specific capabilities. 
In the context of multilingual mathematical reasoning, recent studies~\cite{kang2025multilingual} reveal that this performance gap primarily stems from an \textit{understanding bottleneck}: while LLMs possess strong, language-agnostic reasoning capabilities, their inability to accurately parse and comprehend non-English languages prevents them from fully unlocking this potential.
% We also confirm this phenomenon in instruction models in \S~\ref{sec:bottleneck}.
% In the domain of mathematical reasoning, LLMs can share the language-agnostic reasoning capability across different languages to some extent.

% One promising approach to bridging the gap is multilingual alignment, where LLMs first understand low-source languages through the dominant language, English most of the time, then think in the dominant language, and finally convert the output back into the original language.

A promising strategy to bridge this gap is multilingual alignment~\cite{zhu2024question,she2024mapo}, where LLMs leverage a dominant language (typically English) as a pivot to process lower-resource languages.
However, existing alignment methods face significant limitations.
Implicit alignment approaches~\cite{li-etal-2024-improving-context, bu-etal-2025-alignx}, which attempt to align language representations directly in the latent space, carry a high risk of catastrophic forgetting when applied to post-trained models and lack human-understandable interpretability.
Conversely, training base models with code-switching data during the (continual) pre-training stage~\cite{yoo-etal-2025-code-switching, wang-etal-2025-investigating-scaling} is resource-intensive, and its effectiveness on post-trained models remains unvalidated and risks disrupting the advanced reasoning abilities.

To overcome these challenges, we propose a novel reinforcement learning (RL) approach called \textbf{T}ranslation-\textbf{A}ugmented \textbf{P}olicy \textbf{O}ptimization~(TAPO).
Built upon the on-policy GRPO framework~\cite{shao2024deepseekmath} to mitigate catastrophic forgetting~\cite{shenfeld2025rl}, TAPO enforces an explicit \textit{understand-then-reason} paradigm.
Specifically, the model is trained to first formulate an explicit English translation of the initial problem as a tangible surrogate for ``understanding'', followed by conducting the reasoning steps entirely in English.
Because understanding is manifested as an explicit translation, we can directly quantify and optimize it using standard translation metrics as reward signals.

Crucially, directly combining translation rewards and reasoning rewards within a standard RL trajectory introduces severe reward conflicts.
For instance, a relatively poor translation within the group might still lead to a correct reasoning process, but its resulting negative overall advantage would unfairly penalize the correct reasoning. 
To resolve this, TAPO introduces a \textit{step-level relative advantage} mechanism.
This effectively decouples the credit assignment, calculating advantages for the translation tokens and reasoning tokens completely independently within the same trajectory.
We further compute the final translation advantage as a weighted average of both the translation and reasoning advantages, providing richer task-specific signals to guide the translation process.

% Intuitively, translation into English can be regarded as the model's comprehension of the problem as English is the dominant language.
% Consequently, we can provide rewards of understanding through existing translation metrics to sharpen the model's understanding.
% We regard translation as the model's understanding of the problem, thus we can provide translation rewards through existing translation metrics to sharpen the model's understanding.
% The Translate-Test prompt is used as the system prompt that requires the model to .
% Such kind of explicit alignment by translation, despite not manipulating hidden states directly, may also be capable of aligning the multilingual representations in the middle layers~\cite{zhu2024power}.
% This prevalent algorithm has shown its effectiveness in both mathematical reasoning~\cite{guo2025deepseek} and translation tasks~\cite{feng2025mt}.

Extensive experiments are conducted on both in-domain and out-of-domain multilingual mathematical benchmarks across two different models.
We evaluate not only mathematical accuracy but also the underlying understanding capability by assessing problem translation quality.
Results show that our proposed method successfully synergizes understanding and reasoning capabilities, outperforming baseline methods like naive GRPO. 
Further analysis demonstrates that our step-level reward mechanism successfully avoids severe reward conflicts, and notably, the explicit translation step does not increase the overall inference cost.

The main contributions of our work can be summarized as follows:
\begin{itemize} [nosep]
    \item We propose TAPO, a novel RL framework that enforces an explicit ``understand-then-reason'' paradigm with joint optimization to bridge the multilingual reasoning gap.
    \item We identify and resolve the trajectory-level reward conflict in joint optimization by introducing a step-level relative advantage mechanism, enabling independent credit assignment.
    \item Extensive empirical validations demonstrate that TAPO not only achieves state-of-the-art results on in-domain tasks but also generalizes robustly to unseen languages and out-of-domain mathematical benchmarks.
    % \item Experiments show that our method achieves best performance on both reasoning and translation tasks, and generalizes effectively to untrained languages and out-of-domain tasks.
    % \item Our analyses provide deeper insights into the multilingual alignment and multilingual post-training strategies.
\end{itemize}

\begin{figure}
    \centering
    \includegraphics[width=\linewidth]{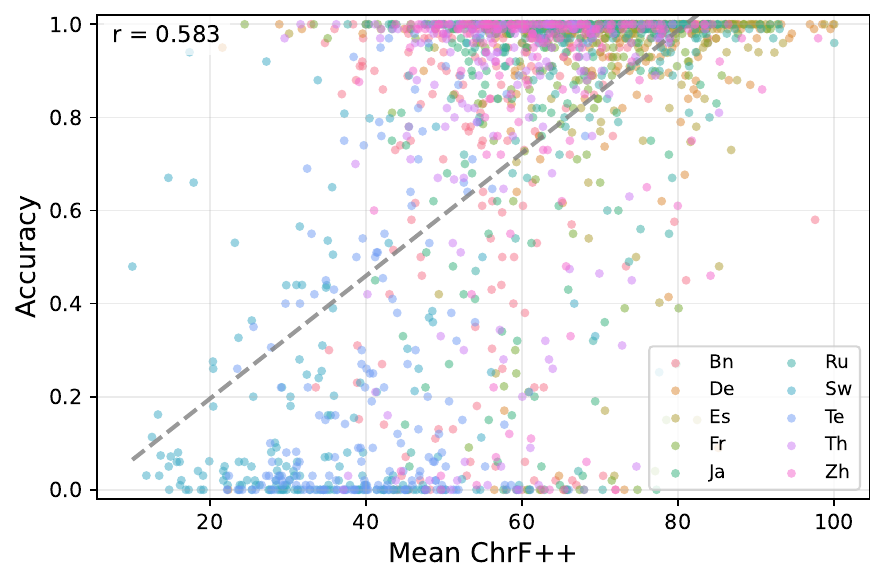}
    \caption{Problem-level relationship between translation quality and reasoning accuracy on MGSM. Each point represents a distinct non-English problem, for which the accuracy of the corresponding English problem is over 0.9. The results are generated by Qwen2.5-3B-Instruct with 100 runs. The x-axis indicates the mean ChrF++ score of the translation, while the y-axis shows the overall accuracy. The dashed line represents the line of best fit, with the Pearson correlation coefficient $r=0.583$, $p<1e-5$.}
    \label{fig:acc_chrfpp}
    \vskip -0.1in
\end{figure}

\begin{figure}[t]
    \centering
    \includegraphics[width=0.85\linewidth]{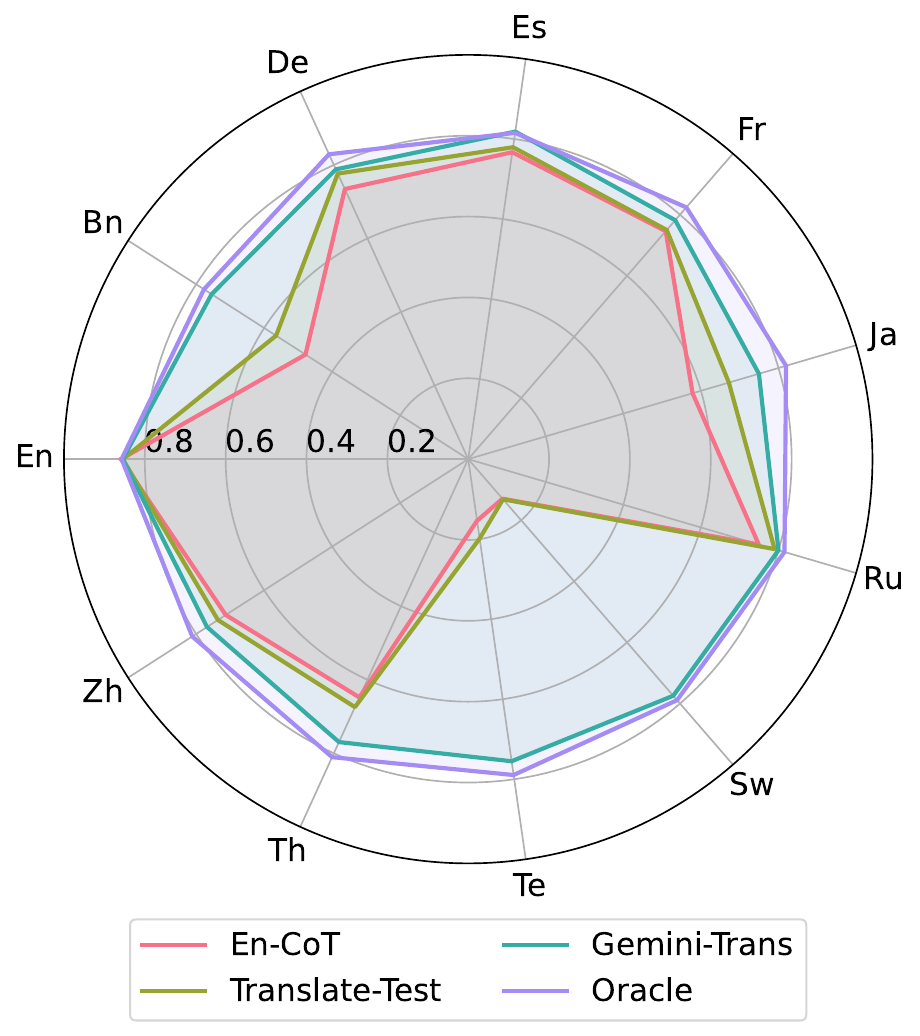}
    \caption{The performance of Qwen2.5-3B-Instruct on MGSM with different prompts. The full prompts are in Appendix~\ref{sec:prompts}. The last two are the same as Translate-Test, but the model's translation is replaced with Gemini2.5-Flash's and the reference, respectively.}
    \label{fig:qwen_mgsm}
    \vskip -0.1in
\end{figure}

\section{Preliminary Studies}
\label{sec:preliminary}

% In this section, we first highlight the primary bottleneck in multilingual reasoning (Section~\ref{sec:bottleneck}) and the limitations of applying Naive GPQA under the direct reasoning mode (Section~\ref{sec:naive_grpo}).

% \subsection{Multilingual Understanding is a Bottleneck}
\subsection{Multilingual Understanding Bottleneck}
\label{sec:bottleneck}
To assess how language comprehension impacts multilingual reasoning, we analyze the correlation between a model's problem understanding and its reasoning accuracy.
We quantify understanding via self-translation quality under a Translate-Test prompt (Table~\ref{tab:transtest}), measured by ChrF++ scores~\cite{popovic-2017-chrf}.
As Figure~\ref{fig:acc_chrfpp} shows, translation quality strongly correlates with accuracy, indicating that bolstering multilingual comprehension is crucial for multilingual reasoning.

We further evaluate the severity of this bottleneck.
Figure~\ref{fig:qwen_mgsm} demonstrates that prompting models to self-translate before reasoning (Translate-Test) yields only marginal gains over direct multilingual-to-English reasoning (En-CoT).
Crucially, this approach significantly underperforms settings using high-quality external translations from superior models (Gemini-Trans) or references (Oracle).
These performance gaps confirm that weak multilingual understanding remains a primary bottleneck for advancing multilingual reasoning.

% Current LLMs still have difficulty fully understanding the multilingual problems.
% We take the quality of the English translations of a problem, ChrF++~\cite{popovic-2017-chrf} used here, as the understanding level of the problem.
% As shown in Figure~\ref{fig:qwen_mgsm}, we evaluate Qwen2.5-3B-Instruct on MGSM~\cite{shi2023language} with different prompts.
% The prompt of Translate-Test, which requires the model to translate a problem to English, brings minor improvement.
% However, when the self-translation is substituted with a high-quality translation, there is a substantial enhancement in performance, with multilingual gaps almost disappearing.
% The reason why the performance in some languages is a bit lower than in English may be due to the influence of the source problems~\cite{huang-etal-2024-lost} or noise.
% Furthermore, we analyze the relationship between the translation quality and the answer correctness.
% As demonstrated in Figure~\ref{fig:bin_acc}, the accuracy improves as ChrF++ scores increase, reflecting a positive correlation between understanding and performance.
% These findings suggest that the multilingual understanding is still a bottleneck for multilingual reasoning, and improving the multilingual understanding is crucial for promoting the alignment.

% \subsection{Naive GRPO Learns Limited Understanding}
\subsection{Limitation of Naive GRPO}
\label{sec:naive_grpo}

A straightforward approach to enhancing multilingual reasoning is to directly apply GRPO~\cite{shao2024deepseekmath} on multilingual data, similar to \citet{huang2025beyond}.
The objective function of GRPO we used in this work is:
\begin{equation}
\begin{split}
&\mathcal{J}_{GRPO}(\theta)=\mathbb{E}_{o_i}\Bigg[\frac{1}{\sum_{i=1}^G|o_i|}\sum_{i=1}^G\sum_{t=1}^{|o_i|}\min\\
&\Big(r_{i,t}\hat{A}_{i,t}, \mathrm{clip}(r_{i,t}, 1-\varepsilon_{l}, 1+\varepsilon_{h}\Big)\hat{A}_{i,t}\Bigg],
\end{split}
\end{equation}
where $r_{i,t}$ denotes the importance sampling ratio, $\hat{A}_{i,t}$ is the advantage, and $o_i$ represents the generated trajectory for a given problem.
% \begin{equation}
% \begin{split}
% r_{i,t}&=\frac{\pi_{\theta}(o_{i,t}|q,o_{i,<t})}{\pi_{\theta_{old}}(o_{i,t}|q,o_{i,<t})},\\
% \hat{A}_{i,t}&=\frac{R_i-\mathrm{mean}(\{R_i\}_{i=1}^G)}{\mathrm{std}(\{R_i\}_{i=1}^G)},
% \end{split}
% \end{equation}
Following DAPO~\cite{yu2025dapo}, we omit the KL divergence loss and incorporate Clip-Higher alongside a Token-Level Policy Gradient Loss.
The clipping thresholds are set to $\varepsilon_l=0.2$ and $\varepsilon_h=0.28$.

We train Qwen2.5-3B-Instruct using GRPO on the Swahili subset of MGSM8KInstruct~\cite{chen-etal-2024-breaking}.
Although reasoning performance improves steadily, reaching 38.95\% from an initial 12.95\% on MGSM-sw, we observe that the model frequently generates unfaithful reasoning traces that coincidentally yield the correct final answer.
As demonstrated in the example below, the model completely misinterprets the semantic context of the prompt, yet correctly extracts the numerical values and their logical relations.
\begin{textbox}
\textbf{Problem:} Joho hutumia komeo 2 za ufumwele wa buluu na nusu ya kiasi hicho cha ufumwele mweupe. Huwa inatumia jumla ya komeo ngapi? (\textit{A robe takes 2 bolts of blue fiber and half that much white fiber. How many bolts in total does it take?}) \\
\textbf{Output:} ... The problem states that there are 2 cages for the birds and half as much space for the pet birds. We need to find the total number of cages used....\textbackslash boxed\{3\}
\end{textbox}
This phenomenon indicates that naive GRPO on multilingual data yields only marginal improvements in true language comprehension.
Consequently, explicit learning signals are necessary to address the understanding deficit.

\section{Methodology}
\label{sec:method}
To address the limitations discussed above, we introduce TAPO.
Our core insight is to explicitly separate problem comprehension from reasoning by introducing a translation-based surrogate reward, coupling it with the standard reasoning reward.
This formulation requires answering two key questions:
\textit{How can we reliably quantify problem understanding?}~(\S~\ref{sec:reward_model}) and \textit{How do we integrate this signal into the RL pipeline?}~(\S~\ref{sec:trans_adv}).

\subsection{Reward Modeling}
\label{sec:reward_model}
Directly quantifying a model's intrinsic understanding of a multilingual problem is challenging.
To construct a reliable learning signal, we use the quality of the model's English self-translation as a surrogate metric for comprehension.
We use the Translate-Test prompt to enforce an understand-then-reason paradigm, translating the question into English prior to generating the reasoning trace.

\paragraph{Format Reward.}
The rollout trajectory $o$ for each sample is the concatenation of the translation snippet $\tau_{trans}$ and the subsequent reasoning trace containing the final answer $\tau_{reason}$, denoted as $o=[\tau_{trans},\tau_{reason}]$.
The output format requires $\tau_{trans}$ to be cleanly enclosed within \texttt{<english\_translation>} tags.
Thus, our format reward is simple:
\begin{equation}
R^{format}=
\begin{cases}
 1,\quad\text{if the translation exists}\\
 0,\quad\text{otherwise}
\end{cases}.
\end{equation}
We do not mandate a CoT process for the translation itself, as prior work shows limited benefits for CoT in direct translation tasks~\cite{wu-etal-2025-please}.

\paragraph{Translation Reward.}
To assess the translation snippet $\tau_{trans}$, we employ established translation metrics~\cite{feng2025mt}.
For n-gram similarity against an English reference $q_{en}$, we use ChrF++~\cite{popovic-2017-chrf}.
For semantic model-based evaluation, we utilize \textsc{xCOMET-XL}~\cite{guerreiro2024xcomet}.
Additionally, we experiment with the reference-free metric \textsc{CometKiwi-DA-XL}~\cite{rei-etal-2023-scaling} to evaluate the necessity of parallel data.
Note that we observe \textsc{xCOMET} is susceptible to reward hacking and struggle with low-resource languages (e.g., Telugu, Swahili).
The total translation reward is masked by the format reward:
% Further, we discover that different languages are suited to evaluation using different metrics.
% For low-resource languages, model-based metrics $M_{model}$ are not more effective than the n-gram based metrics $M_{ngram}$.
% For high-resource languages that models can translate well, model-based metrics can better capture the nuances in translations.
% As a result, we develop an adaptive metric according to the source language.
% In experiments, the adaptive metric use ChrF++ for Swahili and Telugu, and \textsc{xCOMET} for other languages.
\begin{equation}
R^{trans}=M(\tau_{trans},q)\times R^{format},
\end{equation}
where $q$ is the source problem and $M$ is the chosen evaluation metric like ChrF++.

\begin{figure}
    \centering
    \includegraphics[width=\linewidth]{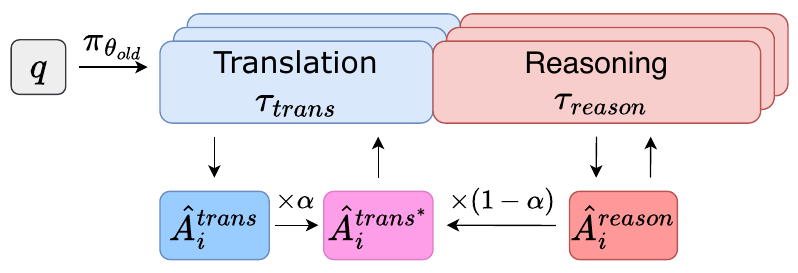}
    \caption{The illustration of the step-level relative advantage calculation in TAPO.}
    \label{fig:tapo}
    \vskip -0.1in
\end{figure}

\paragraph{Reasoning Reward.}
Following standard mathematical reasoning RL setups~\cite{guo2025deepseek}, we utilize exact outcome verification via \texttt{Math-Verify}\footnote{\url{https://github.com/huggingface/Math-Verify}} as our reasoning reward:
% We use the outcome correctness as the reasoning reward, which has proven its effectiveness in mathematics~\cite{guo2025deepseek}.
% We evaluate the correctness by \texttt{Math-Verify}\footnote{\url{https://github.com/huggingface/Math-Verify}}.
\begin{equation}
R^{reason}=
\begin{cases}
 1,\quad\text{if the final answer is correct} \\
 0,\quad\text{otherwise or if } R^{format}=0
\end{cases}.
\end{equation}

\subsection{Step-Level Relative Advantage}
\label{sec:trans_adv}
A naive integration strategy would simply sum the rewards before applying GRPO.
However, this produces severe reward conflicts that hinder reasoning.
For example, if all trajectories in a group achieve the correct answer (or all fail), the final advantage would be entirely dominated by the translation score.
This penalizes valid reasoning traces coupled with mediocre translations, and conversely rewards flawed logic just because the translation was fluent.

To resolve this, we leverage the structural separation explicitly built into the trajectory $o=[\tau_{trans},\tau_{reason}]$ and allocate step-level relative advantages, illustrated in Figure~\ref{fig:tapo}.
Inspired by \citet{zhang2025rlvmr} and \citet{feng2025group}, we normalize the scores independently within their respective segments:
\begin{equation}
\begin{split}
\hat{A}_{i}^{trans}&=\frac{R^{trans}_i-\mathrm{mean}(\{R^{trans}_i\}_{i=1}^G)}{\mathrm{std}(\{R^{trans}_i\}_{i=1}^G)}, \\
\hat{A}_{i}^{reason}&=\frac{R^{reason}_i-\mathrm{mean}(\{R^{reason}_i\}_{i=1}^G)}{\mathrm{std}(\{R^{reason}_i\}_{i=1}^G)}.
\end{split}
\end{equation}
% \begin{equation}
% \hat{A}_{i,t}=
% \begin{cases}
% \frac{R_{trans}^i-\mathrm{mean}(\{R_{trans}^i\}_{i=1}^G)}{\mathrm{std}(\{R_{trans}^i\}_{i=1}^G)}, \,\ \;\text{if $t\in\tau_{trans}$}\\
% \frac{R_{reason}^i-\mathrm{mean}(\{R_{reason}^i\}_{i=1}^G)}{\mathrm{std}(\{R_{reason}^i\}_{i=1}^G)}, \text{if $t\in\tau_{reason}$}
% \end{cases}
% \end{equation}
Here, $\hat{A}_{i}^{trans}$ applies to the tokens in the translation step, and $\hat{A}_{i}^{reason}$ to the reasoning trace tokens.
Finally, to mitigate the reward hacking of translation metrics and couple the translation intent toward accurate problem-solving, we fuse the translation advantage with the reasoning advantage via a simple interpolation coefficient $\alpha \in [0,1]$:
\begin{equation}
\hat{A}^{trans^*}_{i}=\alpha\hat{A}_{i}^{trans}+(1-\alpha)\hat{A}_{i}^{reason}.
\end{equation}

\section{Experiments}
\label{sec:experiments}
\subsection{Experimental Setup}
\paragraph{Base Models.}
We evaluate our method using two instruction-tuned models: Qwen2.5-3B-Instruct~\cite{qwen2.5} and Llama3.2-3B-Instruct~\cite{dubey2024llama}.
For brevity, we refer to them as Qwen and Llama in subsequent sections.

\paragraph{Datasets.}
Our primary training corpus is MGSM8KInstruct~\cite{chen-etal-2024-breaking}.
Additionally, we translate the original GSM8K~\cite{cobbe2021training} training set into Telugu via Gemini2.5-Flash, adopting the prompt from \citet{huang2025beyond}.
We select a subset of five languages to train each model: Bengali~(Bn), Japanese~(Ja), Swahili~(Sw), Telugu~(Te), and Thai~(Th) for Qwen; and Bn, German~(De), Sw, Th, and Chinese~(Zh) for Llama.
This setting yields approximately 7.4k training samples per language.

\paragraph{Benchmarks and Metrics.}
To evaluate multilingual mathematical reasoning, we use MGSM~\cite{shi2023language} as the in-domain test set.
Because MGSM covers eleven languages, the six omitted languages naturally serve as an unseen-language evaluation.
We also include MMATH~\cite{luo2025mmath} and MSVAMP~\cite{chen-etal-2024-breaking} to assess out-of-domain (OOD) reasoning performance.
For robust evaluation, we decode each sample eight times at a temperature of 0.6 and compute the average accuracy.

To evaluate translation capabilities, we leverage the non-English problems from MGSM as source texts and the paired English problems as references.
As with reasoning, each source text is mapped multiple times to compute an average score.
Because ChrF++ and \textsc{xCOMET} are explicitly optimized during our RL training phase, we adopt an LLM-as-a-judge approach using Gemini2.5-Flash to guarantee an unbiased, third-party evaluation~\cite{lavie-etal-2025-findings}.
This metric, denoted as the Gemini Score, prompts the evaluator to assign the translation a definitive score ranging from 0 to 100 (prompt details in Appendix~\ref{sec:prompts}).

\begin{table*}[ht]
    \centering
    \small
    \begin{tabular}{l|cccccc|cccccc|c}
    \toprule
        \multirow{2}{*}{\textbf{Model}} & \multicolumn{5}{c}{\textbf{Trained Languages}} & \multirow{2}{*}{\textbf{Avg}} & \multicolumn{6}{c|}{\textbf{Untrained Languages}} & \multirow{2}{*}{\makecell{\textbf{Total}\\\textbf{Avg}}}\\
         & \textbf{Bn} & \textbf{Ja} & \textbf{Sw} & \textbf{Te} & \textbf{Th} & & \textbf{De} & \textbf{En} & \textbf{Es} & \textbf{Fr} & \textbf{Ru} & \textbf{Zh} & \\ 
        \midrule
        Qwen & 56.5 & 67.2 & 13.2 & 19.9 & 67.5 & 44.8 & 77.7 & 85.5 & 78.0 & 75.0 & 79.0 & 73.6 & 44.8 \\
        + SFT-TransTest & \ \ 2.8 & \ \ 8.8 & \ \ 2.0 & \ \ 1.6 & 10.0 & \ \ 5.0 & 26.0 & 85.6 & 28.0 & 29.6 & 23.2 & 47.2 & 24.1 \\
        + QAlign & 32.0 & 46.4 & \ \ 6.4 & 10.8 & 50.0 & 29.1 & 61.2 & 70.8 & 62.4 & 60.0 & 60.4 & 63.6 & 47.6 \\
        + RAFT & 60.7 & 62.4 & 24.6 & 36.8 & 68.1 & 50.5 & 71.6 & 80.4 & 73.8 & 69.0 & 71.3 & 70.3 & 62.6 \\
        + GRPO-EnCoT & 75.2 & \textbf{78.7} & 39.0 & 51.0 & 82.2 & 65.2 & 84.0 & 89.4 & 84.7 & 82.7 & \textbf{85.8} & 81.4 & 75.8 \\
        + GRPO-TransTest & \textbf{75.7} & 78.2 & 40.9 & 51.2 & \textbf{82.9} & 65.8 & 84.6 & 90.5 & \textbf{87.2} & \textbf{83.9} & 85.4 & 80.6 & 76.4 \\
        \midrule
        + TAPO-CK & 73.6 & 77.3 & 46.7 & \textbf{53.6} & 82.7 & 66.8 & \textbf{85.4} & 89.5 & 85.1 & 83.6 & 84.9 & 79.0 & 76.5 \\
        + TAPO-\textsc{xCOMET} & 75.0 & 78.4 & 45.3 & 51.4 & 82.2 & 66.4 & 83.6 & 89.8 & 84.5 & 82.9 & 84.1 & 80.8 & 76.2 \\
        + TAPO-ChrF++ & 75.0 & 76.8 & \textbf{49.0} & 52.7 & \textbf{82.9} & \textbf{67.3} & 85.0 & \textbf{91.7} & 86.5 & 82.3 & 84.1 & \textbf{82.2} & \textbf{77.1} \\
        \midrule[0.7pt]
        \multirow{2}{*}{\textbf{Model}} & \multicolumn{5}{c}{\textbf{Trained Languages}} & \multirow{2}{*}{\textbf{Avg}} & \multicolumn{6}{c|}{\textbf{Untrained Languages}} & \multirow{2}{*}{\makecell{\textbf{Total}\\\textbf{Avg}}}\\
         & \textbf{Bn} & \textbf{De} & \textbf{Sw} & \textbf{Th} & \textbf{Zh} & & \textbf{En} & \textbf{Es} & \textbf{Fr} & \textbf{Ja} & \textbf{Ru} & \textbf{Te} & \\ 
        \midrule
        Llama & 34.7 & 56.8 & 27.1 & 48.7 & 54.6 & 44.4 & 73.6 & 61.6 & 56.7 & 44.1 & 58.3 & 31.0 & 49.7 \\
        + SFT-TransTest & 45.3 & 45.6 & 49.1 & 44.5 & 50.9 & 47.1 & 55.4 & 64.8 & 59.4 & 54.2 & 54.8 & 53.3 & 52.5 \\
        + QAlign & 30.2 & 29.1 & 31.8 & 27.4 & 38.2 & 31.3 & 43.5 & 57.5 & 47.0 & 44.1 & 43.1 & 41.6 & 39.4 \\
        + RAFT & 62.2 & 73.5 & 64.2 & 66.1 & 69.6 & 67.1 & 81.9 & 76.0 & 73.2 & 64.9 & 73.6 & 59.6 & 69.5 \\
        + GRPO-EnCoT & 66.2 & 76.7 & 68.8 & 71.3 & 73.0 & 71.2 & 86.7 & 79.7 & 76.4 & 69.2 & 76.6 & 60.4 & 73.2 \\
        + GRPO-TransTest & 67.6 & 79.7 & 70.3 & 74.8 & 75.6 & 73.6 & 86.0 & 81.5 & 77.8 & 70.0 & 78.9 & 64.4 & 75.1 \\
        \midrule
        + TAPO-CK & 68.8 & 78.8 & 71.9 & 73.1 & 77.9 & 74.1 & 86.8 & 79.3 & 76.7 & 71.0 & 78.1 & 64.6 & 75.2 \\
        + TAPO-\textsc{xCOMET} & 68.7 & 78.7 & 71.7 & \textbf{76.1} & 77.2 & 74.5 & \textbf{87.6} & 81.4 & 77.6 & 71.4 & \textbf{79.3} & \textbf{66.3} & 76.0 \\
        + TAPO-ChrF++ & \textbf{69.7} & \textbf{80.0} & \textbf{72.1} & 73.8 & \textbf{79.7} & \textbf{75.0} & 87.3 & \textbf{84.6} & \textbf{78.7} & \textbf{72.0} & 78.6 & 64.3 & \textbf{76.4} \\
        \bottomrule
    \end{tabular}
    \vskip -0.05in
    \caption{The performance of each model based on Qwen2.5-3B-Instruct and Llama3.2-3B-Instruct, respectively. The highest scores in each column are in bold. The x in TAPO-x is the translation metric function. CK is short for \textsc{CometKiwi}.}
    \label{tab:mgsm_results}
    \vskip -0.1in
\end{table*}

\paragraph{Baselines.}
\begin{itemize}[nosep]
    \item \textbf{Base}: The original instruction models.
    \item \textbf{SFT-TransTest}: Fine-tuned on the multilingual training data detailed above. The expected output strictly follows the Translate-Test format, generating the English translation concatenated with the original English reasoning trace.
    \item \textbf{QAlign}~\cite{zhu2024question}: Employs a two-stage SFT pipeline comprising a question alignment stage followed by a response alignment stage. We formulate the translation task using our multilingual dataset, and the reasoning task using the original GSM8K dataset.
    \item \textbf{RAFT}~\cite{dong2023raft}: An instantiation of rejection sampling fine-tuning. We use the Translate-Test prompt, retaining only self-generated responses that yield the correct answer to iteratively update the model.
    \item \textbf{GRPO-EnCoT}:  Standard GRPO training applied directly with the En-CoT prompt, exclusively using the reasoning correctness reward.
    \item \textbf{GRPO-TransTest}: Same as GRPO-EnCoT, but trained using the Translate-Test prompt.
\end{itemize}
All SFT and RL baselines are evaluated using their respective training prompts, whereas the Base models directly use the Translate-Test prompt.

\paragraph{Training details.}
We implement our RL pipeline using the \texttt{verl} framework~\cite{sheng2024hybridflow}.
For RL training, the learning rate is fixed at 1e-6, with a global batch size of 256, a mini-batch size of 64, and a group size of 8.
During rollout, the maximum response length is constrained to 2048 tokens and the temperature is set to 1.0.
We train Qwen for 5 epochs and Llama for 2 epochs.
The hyper-parameter $\alpha$ is set to 0.25 across all models.
For all SFT baselines, we set the learning rate to 1e-6 and the global batch size to 128.

\subsection{Main Results}
\paragraph{TAPO demonstrates superior performance on trained languages and comparable performance on untrained languages.}
Table~\ref{tab:mgsm_results} presents the reasoning accuracy on MGSM.
For trained languages, Qwen optimized with TAPO-ChrF++ achieves the best average scores.
The most significant gains are observed in the two low-resource languages, Swahili and Telugu, which outperform the GRPO-TransTest baseline by 8.1\% and 1.5\%, respectively. Furthermore, its performance on untrained languages remains highly competitive with standard GRPO.
For the Llama model, TAPO-ChrF++ yields even more substantial and uniform performance improvements across both seen and unseen languages.
Overall, the significant effectiveness on trained languages confirms that explicitly bolstering translation comprehension inherently benefits downstream reasoning, particularly for low-resource languages where the understanding bottleneck is most severe.

\begin{table*}[ht]
    \centering
    \small
    \begin{tabular}{l|ccccccccccc}
    \toprule
        \textbf{Model} & \textbf{Bn} & \textbf{De} & \textbf{Es} & \textbf{Fr} & \textbf{Ja} & \textbf{Ru} & \textbf{Sw} & \textbf{Te} & \textbf{Th} & \textbf{Zh} & \textbf{Avg} \\
        \midrule
        Qwen & 58.60 & 84.07 & 87.30 & 85.89 & 81.10 & 87.28 & \ \ 7.72 & 17.82 & 77.68 & \textbf{86.72} & 67.42 \\
        + GRPO-EnCoT & 62.97 & 83.58 & 87.88 & 86.66 & 80.26 & \textbf{88.88} & 18.18 & 29.31 & 77.99 & 83.66 & 69.94 \\
        + GRPO-TransTest & 62.59 & 82.69 & 86.76 & 85.89 & 79.91 & 86.85 & 19.94 & 33.67 & 79.09 & 83.87 & 70.12 \\
        + TAPO-CK & 67.06 & 85.82 & 89.02 & \textbf{87.21} & 82.48 & 87.51 & \textbf{30.03} & \textbf{42.90} & 80.96 & 85.67 & 73.87 \\
        + TAPO-\textsc{xCOMET} & \textbf{70.88} & 51.49 & 51.52 & 54.85 & \textbf{84.47} & 87.07 & \ \ 0.00 & 41.64 & \textbf{82.71} & 85.72 & 61.04 \\
        + TAPO-ChrF++ & 68.89 & \textbf{87.47} & \textbf{89.83} & 87.11 & 83.32 & 88.80 & 29.29 & 42.37 & 82.70 & 86.60 & \textbf{74.64} \\
        \midrule
        Llama & 59.59 & 74.88 & 81.14 & 76.92 & 59.90 & 76.75 & 50.17 & 55.55 & 59.44 & 65.67 & 66.00 \\
        + GRPO-EnCoT & 56.90 & 64.54 & 71.26 & 66.40 & 53.91 & 64.92 & 47.86 & 54.32 & 54.96 & 57.92 & 59.30 \\
        + GRPO-TransTest & 66.38 & 78.76 & 81.44 & 81.30 & 69.55 & 78.35 & 54.99 & 62.74 & 70.76 & 75.58 & 71.98 \\
        + TAPO-CK & 72.12 & 83.16 & 87.62 & 84.38 & 75.92 & 83.76 & 59.50 & 68.00 & 76.44 & 81.76 & 77.27 \\
        + TAPO-\textsc{xCOMET} & 72.52 & 84.60 & 87.86 & 85.30 & 74.71 & 84.28 & 61.84 & \textbf{68.49} & 76.13 & 81.87 & 77.76 \\
        + TAPO-ChrF++ & \textbf{72.89} & \textbf{85.87} & \textbf{89.25} & \textbf{85.60} & \textbf{75.96} & \textbf{85.77} & \textbf{63.62} & 68.28 & \textbf{77.30} & \textbf{83.71} & \textbf{78.82} \\
    \bottomrule
    \end{tabular}
    \caption{The Gemini Scores of each model on the task of translating the MGSM problems from non-English to English. The score of each sample is averaged by eight random runs.}
    \label{tab:trans_results}
    \vskip -0.1in
\end{table*}

\paragraph{TAPO explicitly enhances translation quality, signaling improved problem comprehension.}
We evaluate problem translation primarily using the Gemini Score (with ChrF++ and BLEU~\cite{papineni2002bleu} scores provided in Appendix~\ref{sec:detail_results}).
Translations are either extracted from the reasoning trajectory generated by the Translate-Test prompt or obtained via a separate translation prompt.
As reported in Table~\ref{tab:trans_results}, our method substantially enhances translation quality across nearly all languages, including zero-shot untrained languages.
Specifically, models trained with TAPO-ChrF++ consistently achieve the highest average scores, significantly outperforming the GRPO baselines.
One exception is the Qwen model trained with TAPO-\textsc{xCOMET}, which exhibits performance degradation in languages like De, Es, Fr, and Sw.
This occurs due to reward hacking, where the model exploits the metric's fragility by simply copying the source text.
Interestingly, despite receiving no direct translation rewards, GRPO-TransTest slightly improves translation quality.
This suggests that the downstream reasoning reward provides implicit, albeit weak, optimization signals for the upstream translation, further validating our design of coupling the reasoning and translation advantages.

\begin{table}[t]
    \centering
    \small
    \begin{tabular}{l|cc}
    \toprule
        \textbf{Model} & \textbf{MMATH} & \textbf{MSVAMP} \\
        \midrule
        Qwen & 54.25 & 77.55 \\
        + GRPO-EnCoT & 55.85 & 77.28 \\
        + GRPO-TransTest & \textbf{57.63} & 76.39 \\
        + TAPO-ChrF++ & 56.77 & \textbf{80.21} \\
        \midrule
        Llama & 28.07 & 61.43 \\
        + GRPO-EnCoT & 42.22 & 73.75 \\
        + GRPO-TransTest & 42.34 & 76.02 \\
        + TAPO-ChrF++ & \textbf{43.78} & \textbf{77.25} \\
    \bottomrule
    \end{tabular}
    \caption{The performance of each model on OOD tasks. The scores are averaged across all the languages in the benchmarks.}
    \label{tab:ood_results}
    \vskip -0.1in
\end{table}

\paragraph{TAPO's generalizability extends to OOD reasoning tasks.}
We evaluate the best-performing model variant, TAPO-ChrF++, on OOD tasks.
Detailed results are available in Appendix~\ref{sec:detail_results}.
As shown in Table~\ref{tab:ood_results}, TAPO yields consistent improvements across most tasks, whereas the GRPO-TransTest baseline for Qwen occasionally exhibits performance degradation (e.g., on MSVAMP).
While TAPO's improvements on MMATH are relatively modest, this is likely because the majority of languages in that benchmark were unseen during training, corroborating our findings regarding the difficulty of cross-lingual transfer without explicit alignment data.

\paragraph{SFT induces catastrophic forgetting in post-trained models.}
The results for SFT-TransTest and QAlign (Table~\ref{tab:mgsm_results}) reveal a severe degradation in the reasoning capabilities of the base models across many languages, particularly in high-resource ones.
Strikingly, Qwen's performance drops dramatically after naive SFT, indicating severe catastrophic forgetting.
Conversely, on-policy methods, including RAFT, GRPO, and TAPO, effectively avert catastrophic forgetting, ensuring consistent, monotonic enhancements to multilingual reasoning.

\section{Analysis}
\label{sec:analysis}
In this section, we provide detailed ablation studies and further analyses to better understand the mechanisms and implications underlying TAPO.

\begin{table}[ht]
    \centering
    \small
    \begin{tabular}{l|ccc}
    \toprule
        \textbf{Type} & \textbf{MGSM} & \textbf{MMATH} & \textbf{MSVAMP} \\
        \midrule
        \rowcolor{lightgray!30} \textit{Qwen} & & & \\
        Traj-level & 75.45 & 56.37 & 78.35 \\
        Step-level & \textbf{77.08} & \textbf{56.77} & \textbf{80.21} \\
        \midrule
        \rowcolor{lightgray!30} \textit{Llama} & & & \\
        Traj-level & 74.11 & 42.13 & 75.87 \\
        Step-level & \textbf{76.41} & \textbf{43.78} & \textbf{77.25} \\
    \bottomrule
    \end{tabular}
    \caption{The comparison between the trajectory-level reward by addition and our step-level reward. Scores are averaged across all languages.}
    \label{tab:traj_reward}
\end{table}

\begin{figure}[ht]
    \centering
    \includegraphics[width=\linewidth]{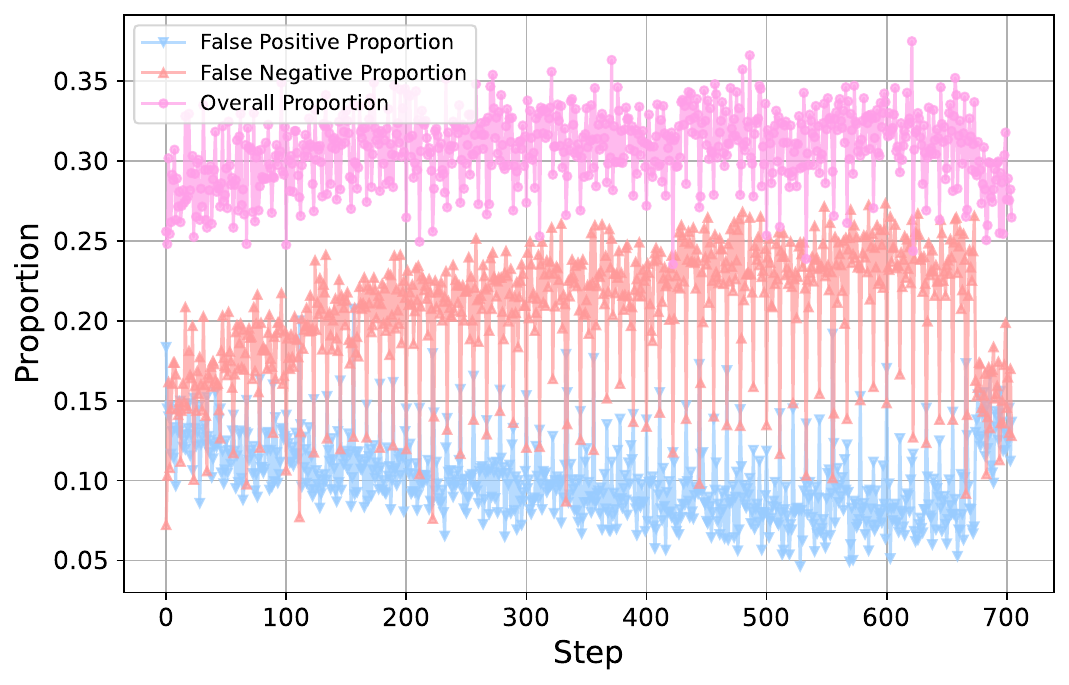}
    \caption{The proportions of false positive and false negative advantages across each training step. The overall proportion is the sum of the two proportions.}
    \label{fig:reward_conflict}
    \vskip -0.1in
\end{figure}

\subsection{Reward Conflict in the Trajectory-Level Reward}
We first compare our step-level reward mechanism against a standard trajectory-level reward, which directly applies the sum $R^{trans}+R^{reason}$ to the entire trajectory.
The outcomes detailed in Table~\ref{tab:traj_reward} demonstrate the consistent superiority of the step-level reward over the trajectory-level approach across different models and evaluation benchmarks.
Furthermore, we identify a severe reward conflict problem introduced by trajectory-level rewards, which fundamentally hinders training effectiveness.
Specifically, advantages derived from normalized trajectory-level rewards can take incorrect signs if the translation reward heavily dominates the total signal.
An advantage assigned to a trajectory is considered a false positive if it is positive despite an incorrect final answer, and a false negative if it is negative despite a correct answer.
As illustrated in Figure~\ref{fig:reward_conflict}, we tracked the proportions of these false positive and false negative advantages throughout the training steps of the Qwen model using the trajectory-level ChrF++ reward. 
The total conflict proportion hovers around 30\%, which introduces significant noise and exerts a detrimental influence on the policy's learning trajectory.

\subsection{Inference Cost}
\begin{figure}
    \centering
    \includegraphics[width=\linewidth]{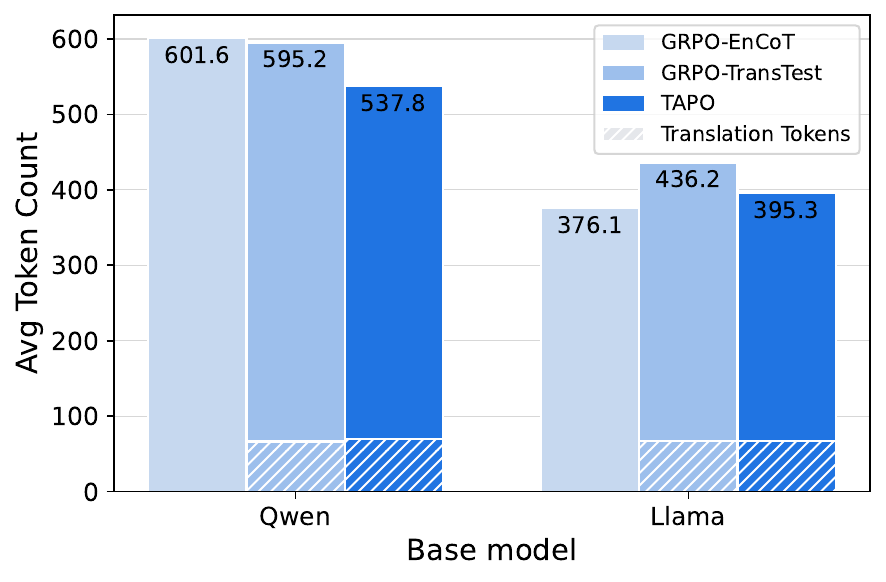}
    \caption{The average number of response tokens on MGSM across all languages. Each bar represents total number of tokens of a response, and the hatched area represents the number of translation tokens.}
    \label{fig:inference_cost}
\end{figure}

To determine whether the explicit translation step imposes significant computational overhead, we evaluate the number of generated tokens on the MGSM benchmark across all tested languages.
As depicted in Figure~\ref{fig:inference_cost}, for the Qwen models, the average token count produced by TAPO is surprisingly lower than that of the standard GRPO variants.
The number of translation tokens generated remains remarkably consistent across different models at approximately 70 tokens, accounting for less than 13.0\% of Qwen's total response length.
Similarly, for Llama models, TAPO still yields shorter overall outputs compared to GRPO-TransTest.
It is important to note that the MGSM benchmark consists of grade-school-level math problems, which generally do not require exceptionally long reasoning chains.
For more complex problems requiring extended analytical steps, the proportion of inference compute allocated to translation tokens would become even more negligible.

\subsection{Impacts of Translation Metrics}
The choice of the underlying translation metric has varying degrees of impact depending on the target language.
As shown in Table~\ref{tab:mgsm_results}, the string-based metric ChrF++ broadly enhances reasoning performance across multiple languages, proving particularly instrumental for low-resource languages like Sw.
In contrast, the model-based metric \textsc{xCOMET} is much more susceptible to reward hacking during RL training.
The policy model quickly learns an exploit: merely copying the non-English source text verbatim often yields an artificially high score, severely degrading the intended learning signal. 
This dynamic contributes to the inferior translation quality observed in the Qwen model when trained with TAPO-\textsc{xCOMET} on certain languages (see Table~\ref{tab:trans_results}).
Conversely, the reference-free metric \textsc{CometKiwi} exhibits greater robustness against reward hacking and achieves reasoning performance comparable to ChrF++. 
These results suggest that, rather than necessitating parallel pairs, TAPO depends on the use of high-quality, robust translation metrics.

\subsection{Influence of the Hyper-parameter $\alpha$}
\begin{figure}[t]
    \centering
    \includegraphics[width=\linewidth]{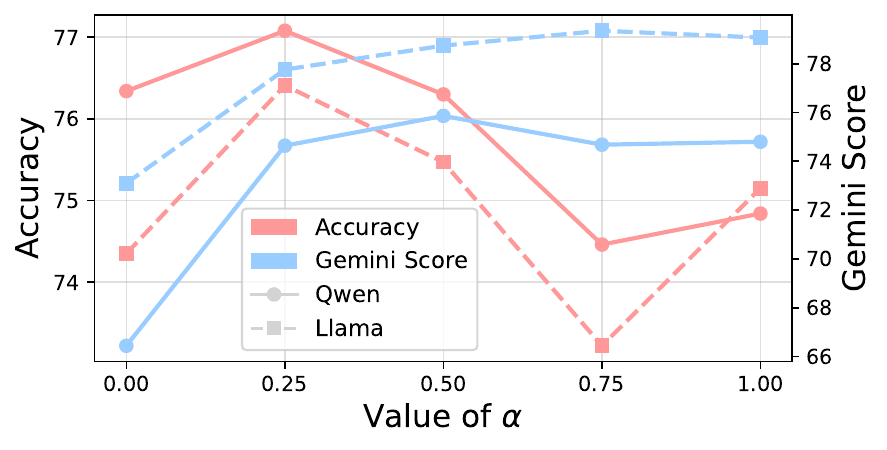}
    \caption{Accuracy and Gemini Score trends for Qwen and Llama models trained by TAPO-ChrF++ across different $\alpha$ values.}
    \label{fig:alpha}
    \vskip -0.1in
\end{figure}

The hyper-parameter $\alpha$ dictates the fusion weight between the translation metric advantage and the reasoning correctness advantage.
As illustrated in Figure~\ref{fig:alpha}, adjusting $\alpha$ reveals a clear trade-off between reasoning accuracy and translation fidelity.
Reasoning accuracy demonstrates a non-monotonic trend: it achieves a global peak at $\alpha=0.25$, then declines sharply to a minimum at $\alpha=0.75$.
Conversely, the translation quality sees a steep improvement from 0 to 0.25, and plateaus with minor fluctuations when $\alpha$ climbs above 0.25.
These behavioral patterns are highly consistent across both models we evaluated.
We speculate that a small translation signal is enough to improve problem understanding and reasoning, while an overly large signal risks drowning out the logical constraints needed for solving problems.

\subsection{Multilingual Learning Dynamics}
\begin{figure}[t]
    \centering
    \includegraphics[width=\linewidth]{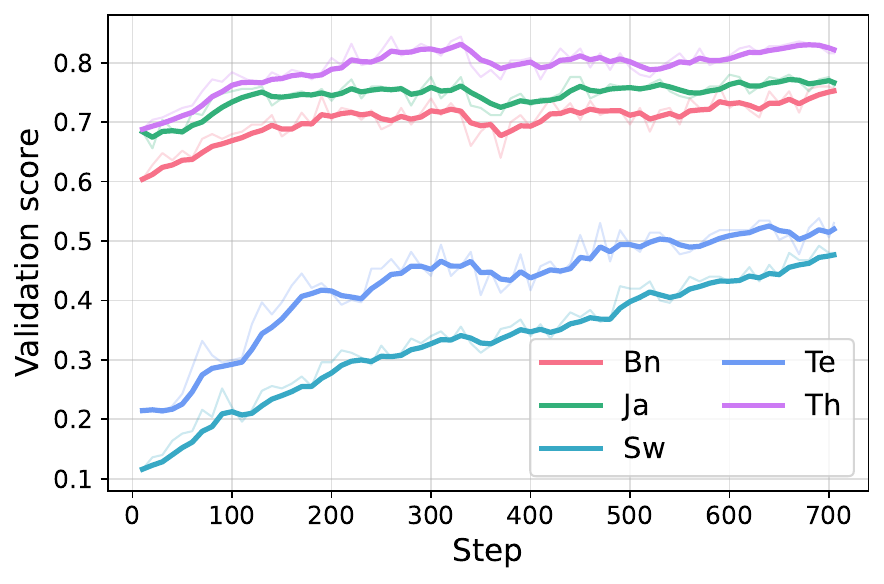}
    \caption{The validation scores of Qwen + TAPO-ChrF++. The bold lines are smoothed by Exponential Moving Average.}
    \label{fig:converge}
    \vskip -0.1in
\end{figure}

By tracking validation performance on MGSM, we observe varied convergence speeds among different languages.
As illustrated in Figure~\ref{fig:converge}, performance on the three mid-resource languages converges early in the training process. 
Notably, the evaluation score for Ja achieves a plateau around step 300 and even exhibits slight degradation in subsequent training steps.
In stark contrast, scores for the two low-resource languages, Sw and Te, exhibit a continuous upward trajectory until the completion of training, showing no signs of saturation.
These learning dynamics suggest that blindly optimizing converged languages provides diminishing, or even negative, returns.
Future improvements could involve adaptively allocating more computational resources or implementing dynamic sampling strategies~\cite{yu2025dapo} tailored toward low-resource languages, which we leave to future work.

\section{Related Works}
\label{sec:rw}
\paragraph{Multilingual Reasoning.}

% potential
% \cite{Multilingual_Transfer_Ability,How_Much_Is_Needed}

% % understanding in reasoning
% \cite{zhu2024question, Getting_More_from_Less}

% \cite{Consistency_RL, Consistency_RL_on_MT_question}

% % potential of first understanding
% \cite{Cross_lingual_Collapse, Multilingual_Paths_Affect}

% % understanding then reasoning
% \cite{huang-etal-2023-languages}

Despite the dominant performance in English, prior studies demonstrate that LLMs retain substantial potential for multilingual reasoning~\cite{Multilingual_Transfer_Ability, How_Much_Is_Needed}.
To exploit this potential, existing methods primarily enhance cross-lingual alignment through additional SFT~\cite{Getting_More_from_Less} or RL~\cite{she2024mapo, Consistency_RL, Consistency_RL_on_MT_question}, aiming to improve multilingual reasoning while preserving established reasoning behaviors.
However, recent analyses show that LLMs exhibit significantly stronger semantic understanding in English than in non-English languages~\cite{Cross_lingual_Collapse, Multilingual_Paths_Affect}, suggesting that reasoning directly in non-English impairs performance. 

In contrast to prior approaches that rely on reasoning natively, we adopt an understand-then-reason paradigm, expressing non-English inputs in English to reduce comprehension difficulty before reasoning.
\citet{huang-etal-2023-languages} encourage translation before reasoning via prompting, but limited by the model’s intrinsic translation ability.
Sharing the similar motivation, QAlign~\cite{zhu2024question} strengthens multilingual understanding by SFT with question translation, followed by English reasoning enhancement.
However, this two-stage paradigm can lead to catastrophic forgetting in the post-trained models, and cannot leverage the synergy between understanding and reasoning.
In contrast, we utilize on-policy RL algorithm and optimize understanding and reasoning jointly.
% In contrast, we explicitly train translation and reasoning with reward design.

\paragraph{Multilingual Alignment.}

% Code Switch
% \cite{yoo-etal-2025-code-switching, wang-etal-2025-investigating-scaling}

% % contrastive learning
% \cite{PreAlign, bu-etal-2025-alignx, li-etal-2024-improving-context}

% % latent disentangle
% \cite{zhao2025less,CC_Tuning}

% % model merge
% \cite{langbridge,mindmerger}

% % RL
% \cite{huang2025beyond, Korean_RL}

To reduce the multilingual disparity, a variety of alignment methods have been proposed in recent years.
Some approaches inject multilingual knowledge by training on code-switched data~\cite{yoo-etal-2025-code-switching, wang-etal-2025-investigating-scaling}, while others employ contrastive learning to map representations into a shared semantic space~\cite{PreAlign, bu-etal-2025-alignx, li-etal-2024-improving-context}.
Additional works activate language-related parameters through latent disentanglement~\cite{zhao2025less,CC_Tuning}, or incorporate external models to assist multilingual understanding~\cite{langbridge,mindmerger}. 
These methods effectively achieve multilingual representation alignment, while offering limited attention to the enhancement of reasoning behaviors. 

RL has been widely validated as an effective approach for enhancing reasoning capabilities in LLMs~\cite{shao2024deepseekmath,guo2025deepseek}, and its benefits extend to multilingual settings as well~\cite{huang2025beyond, Korean_RL, she2024mapo}.
However, existing RL-based approaches largely overlook the distinction between multilingual understanding and reasoning processes.
To address the limitation, we propose an RL framework that jointly optimizes multilingual understanding and reasoning, further unlocking the potential of LLMs for multilingual reasoning.

\section{Conclusion}
\label{sec:conclusion}
In this paper, we introduced TAPO to overcome the multilingual reasoning bottleneck in LLMs. 
TAPO enforces an explicit ``understand-then-reason'' paradigm by translating problems into English before reasoning. 
To resolve reward conflicts during joint optimization, we employ a step-level relative advantage mechanism to independently evaluate translation quality and reasoning correctness. 
Extensive experiments demonstrate that TAPO achieves superior performance in multilingual reasoning and translation, while generalizing robustly to untrained languages and out-of-domain benchmarks.

\section*{Limitations}
The limitations of our work are discussed below:
\begin{itemize}[nosep]
    \item We rely on relatively efficient translation metrics due to computational and cost constraints, foregoing more advanced but expensive LLM-as-a-judge approaches with frontier models. Given that the choice of metric significantly impacts our framework, more robust evaluators could yield further gains.
    \item The reasoning trace and the final answer are currently generated in English, which may be unhelpful for non-English speakers. Additionally, translating all problems into English might discard subtle cultural contexts that are difficult to convey, potentially causing a loss of essential information before reasoning begins. However, in many scenarios, particularly for language-agnostic domains like mathematics, the accuracy of the answer remains the primary objective.
    \item We restrict our evaluation to smaller models and did not extend our training to larger models that inherently possess stronger multilingual capabilities. Furthermore, we did not apply this method to large reasoning models, which may exhibit distinct response dynamics under such alignment strategies.
\end{itemize}

% \section*{Acknowledgments}

% This document has been adapted
% by Steven Bethard, Ryan Cotterell and Rui Yan
% from the instructions for earlier ACL and NAACL proceedings, including those for
% ACL 2019 by Douwe Kiela and Ivan Vuli\'{c},
% NAACL 2019 by Stephanie Lukin and Alla Roskovskaya,
% ACL 2018 by Shay Cohen, Kevin Gimpel, and Wei Lu,
% NAACL 2018 by Margaret Mitchell and Stephanie Lukin,
% Bib\TeX{} suggestions for (NA)ACL 2017/2018 from Jason Eisner,
% ACL 2017 by Dan Gildea and Min-Yen Kan,
% NAACL 2017 by Margaret Mitchell,
% ACL 2012 by Maggie Li and Michael White,
% ACL 2010 by Jing-Shin Chang and Philipp Koehn,
% ACL 2008 by Johanna D. Moore, Simone Teufel, James Allan, and Sadaoki Furui,
% ACL 2005 by Hwee Tou Ng and Kemal Oflazer,
% ACL 2002 by Eugene Charniak and Dekang Lin,
% and earlier ACL and EACL formats written by several people, including
% John Chen, Henry S. Thompson and Donald Walker.
% Additional elements were taken from the formatting instructions of the \emph{International Joint Conference on Artificial Intelligence} and the \emph{Conference on Computer Vision and Pattern Recognition}.

% Bibliography entries for the entire Anthology, followed by custom entries
%\bibliography{custom,anthology-overleaf-1,anthology-overleaf-2}

% Custom bibliography entries only
\bibliography{custom}

\appendix

\newpage
\clearpage

\section{Prompts}
\label{sec:prompts}

\begin{table}[t]
\centering
\begin{promptblock}
[System]
Please solve the following problem by providing a detailed, step-by-step response.

Your output **must** be structured into the following three sections:

### 1. Translation

* Translate the provided problem into English.
* If the problem is already in English, simply reproduce the original problem statement.
* Enclose the final English text within `<english_translation>` and `</english_translation>` tags.

### 2. Thought

* Present your comprehensive, step-by-step reasoning process for arriving at the solution.
* This reasoning must be transparent, logical, and easy to follow. Break down your process to cover problem analysis, strategy & Planning, step-by-step execution, and verification.

### 3. Solution

* After reasoning, provide the final, clear, and accurate answer.
* If the result is a closed-form answer (such as a numerical value, formula, or multiple-choice selection), please enclose it using the `\boxed{}` format.

[User]
{question}
\end{promptblock}
\caption{The prompt of Translate-Test.}
\label{tab:transtest}
\end{table}

We present the prompts used in this section.
Table~\ref{tab:transtest} and Table~\ref{tab:encot} show the Translate-Test prompt and the En-CoT prompt.
Table~\ref{tab:translation_eval} is the prompt used to evaluate the translation quality.

\begin{table}[h]
\centering
\begin{promptblock}
[System]
Please solve the following problem by providing a detailed, step-by-step response.

Your output **must** be structured into the following two sections:

### 1. Thought

* Present your comprehensive, step-by-step reasoning process in English for arriving at the solution.
* This reasoning must be transparent, logical, and easy to follow. Break down your process to cover problem analysis, strategy & Planning, step-by-step execution, and verification.

### 2. Solution

* After reasoning, provide the final, clear, and accurate answer.
* If the result is a closed-form answer (such as a numerical value, formula, or multiple-choice selection), please enclose it using the `\boxed{}` format.

[User]
{question}
\end{promptblock}
\caption{The prompt of En-CoT}
\label{tab:encot}
\end{table}

\begin{table}[t]
\centering
\begin{promptblock}
Score the following translation from {source_lang} to {target_lang} on a scale from 0 to 100, where a score of 0 means a broken or poor translation; 33 indicates a flawed translation with significant issues; 66 indicates a good translation with only minor issues in grammar, fluency, or consistency; and 100 represents a perfect translation in both meaning and grammar.

Answer with only a whole number representing the score, and nothing else.

{source_lang} source text: {source_seg}
{target_lang} translation: {target_seg}
{target_lang} reference: {reference_seg}
\end{promptblock}
\caption{The prompt of LLM-as-a-judge for translation evaluation.}
\label{tab:translation_eval}
\end{table}

\section{The Detailed Experimental Results}
\label{sec:detail_results}
Table~\ref{tab:chrf_results} shows the detailed ChrF++ scores, and Table~\ref{tab:bleu_results} shows the BLEU~\cite{papineni2002bleu} scores.
The model ranks are generally consistent with those scored by Gemini in Table~\ref{tab:trans_results}.

Table~\ref{tab:mmath_results} and Table~\ref{tab:msvamp_results} demonstrate the performance in each language on the OOD tasks.
On MMATH~\cite{luo2025mmath}, most languages are not trained during our RL process, except Ja and Th for Qwen, and Th, Zh for Llama.
On MSVAMP~\cite{chen-etal-2024-breaking}, about half of languages are trained.
TAPO-ChrF++ outperforms baselines on almost all languages, demonstrating the generalization of our method.

\begin{table*}[t]
    \centering
    \small
    \begin{tabular}{l|ccccccccccc}
    \toprule
        \textbf{Model} & \textbf{Bn} & \textbf{De} & \textbf{Es} & \textbf{Fr} & \textbf{Ja} & \textbf{Ru} & \textbf{Sw} & \textbf{Te} & \textbf{Th} & \textbf{Zh} & \textbf{Avg} \\
        \midrule
        Qwen & 55.57 & 70.06 & 72.87 & 65.20 & 59.39 & 67.73 & 33.69 & 40.68 & 57.97 & 62.00 & 58.52 \\
        + GRPO-EnCot & 56.85 & 70.49 & 73.00 & 64.97 & 59.07 & 68.13 & 41.57 & 44.16 & 58.50 & 60.14 & 59.69 \\
        + GRPO-TransTest & 54.76 & 68.84 & 71.25 & 63.42 & 56.96 & 66.42 & 41.22 & 44.32 & 56.99 & 59.27 & 58.35 \\
        + TAPO-CK & 58.56 & 70.65 & 72.77 & 64.54 & 59.38 & 66.92 & 47.18 & 50.59 & 59.30 & 60.61 & 61.05 \\
        + TAPO-\textsc{xCOMET} & 58.11 & 48.54 & 49.98 & 47.86 & 61.22 & 67.41 & 15.68 & 49.36 & 60.12 & 60.52 & 51.88 \\
        + TAPO-ChrF++ & \textbf{63.93} & \textbf{74.34} & \textbf{75.83} & \textbf{67.49} & \textbf{64.67} & \textbf{71.48} & \textbf{51.64} & \textbf{56.87} & \textbf{63.71} & \textbf{64.68} & \textbf{65.46} \\
        \midrule
        Llama & 53.78 & 66.67 & 70.09 & 60.77 & 50.86 & 63.12 & 54.74 & 52.99 & 49.69 & 51.07 & 57.38 \\
        + GRPO-EnCot & 48.98 & 55.66 & 59.71 & 50.72 & 43.81 & 50.51 & 47.84 & 48.51 & 43.62 & 40.33 & 48.97 \\
        + GRPO-TransTest & 54.91 & 67.38 & 69.71 & 61.35 & 53.42 & 62.22 & 55.40 & 54.77 & 52.67 & 53.42 & 58.53 \\
        + TAPO-CK & 60.16 & 70.83 & 72.70 & 63.26 & 57.06 & 65.92 & 60.04 & 59.73 & 57.81 & 58.45 & 62.60 \\
        + TAPO-\textsc{xCOMET} & 60.33 & 71.31 & 72.54 & 64.57 & 57.91 & 66.57 & 59.51 & 59.97 & 58.29 & 59.64 & 63.06 \\
        + TAPO-ChrF++ & \textbf{64.72} & \textbf{75.37} & \textbf{76.29} & \textbf{66.70} & \textbf{61.73} & \textbf{70.37} & \textbf{65.24} & \textbf{64.52} & \textbf{62.49} & \textbf{64.21} & \textbf{67.16} \\
    \bottomrule
    \end{tabular}
    \caption{The ChrF++ scores of each model on the task of translating the MGSM problems from non-English to English. The score of each sample is averaged by eight random runs.}
    \label{tab:chrf_results}
\end{table*}

\begin{table*}[h]
    \centering
    \small
    \begin{tabular}{l|ccccccccccc}
    \toprule
        \textbf{Model} & \textbf{Bn} & \textbf{De} & \textbf{Es} & \textbf{Fr} & \textbf{Ja} & \textbf{Ru} & \textbf{Sw} & \textbf{Te} & \textbf{Th} & \textbf{Zh} & \textbf{Avg} \\
        \midrule
        Qwen & 33.75 & 52.92 & 57.28 & 46.55 & 37.92 & 50.01 & 14.18 & 18.26 & 34.33 & 41.19 & 38.64 \\
        + GRPO-EnCot & 35.04 & 53.48 & 57.67 & 46.16 & 37.77 & 50.51 & 20.61 & 22.00 & 34.10 & 39.37 & 39.67 \\
        + GRPO-TransTest & 34.28 & 51.78 & 55.49 & 44.32 & 35.81 & 48.81 & 21.54 & 21.70 & 35.31 & 38.29 & 38.73 \\
        + TAPO-CK & 38.57 & 53.49 & 57.58 & 45.51 & 38.47 & 48.63 & 27.70 & 29.54 & 38.05 & 39.83 & 41.74 \\
        + TAPO-\textsc{xCOMET} & 34.37 & 31.99 & 34.10 & 28.79 & 40.41 & 48.66 & 1.83 & 26.17 & 37.08 & 40.04 & 32.35 \\
        + TAPO-ChrF++ & \textbf{42.08} & \textbf{58.41} & \textbf{60.88} & \textbf{49.49} & \textbf{43.98} & \textbf{54.33} & \textbf{30.71} & \textbf{33.84} & \textbf{40.51} & \textbf{44.65} & \textbf{45.89} \\
        \midrule
        Llama & 32.04 & 49.18 & 54.73 & 41.48 & 28.78 & 44.63 & 35.44 & 31.41 & 27.03 & 29.44 & 37.41 \\
        + GRPO-EnCot & 19.08 & 22.84 & 28.62 & 20.72 & 14.06 & 19.22 & 19.40 & 18.75 & 14.30 & 11.88 & 18.89 \\
        + GRPO-TransTest & 32.21 & 48.66 & 52.90 & 40.91 & 30.29 & 41.54 & 35.74 & 31.79 & 28.76 & 30.10 & 37.29 \\
        + TAPO-CK & 39.93 & 53.49 & 57.78 & 43.60 & 35.87 & 47.01 & 42.68 & 39.21 & 36.84 & 36.99 & 43.34 \\
        + TAPO-\textsc{xCOMET} & 39.16 & 54.88 & 57.01 & 45.56 & 36.19 & 48.27 & 41.19 & 39.18 & 36.49 & 38.34 & 43.63 \\
        + TAPO-ChrF++ & \textbf{43.27} & \textbf{60.55} & \textbf{62.09} & \textbf{48.86} & \textbf{41.28} & \textbf{53.06} & \textbf{48.13} & \textbf{42.79} & \textbf{39.66} & \textbf{43.00} & \textbf{48.27} \\
    \bottomrule
    \end{tabular}
    \caption{The BLEU scores of each model on the task of translating the MGSM problems from non-English to English. The score of each sample is averaged by eight random runs.}
    \label{tab:bleu_results}
\end{table*}

\begin{table*}[h]
    \centering
    \small
    \begin{tabular}{l|ccccccccccc}
    \toprule
        \textbf{Model} & \textbf{Ar} & \textbf{En} & \textbf{Es} & \textbf{Fr} & \textbf{Ja} & \textbf{Ko} & \textbf{Pt} & \textbf{Th} & \textbf{Vi} & \textbf{Zh} & \textbf{Avg} \\
    \midrule
        Qwen & 52.07 & 56.78 & 56.32 & 56.32 & 52.01 & 52.71 & 56.35 & 51.97 & 52.67 & 55.31 & 54.25 \\
        + GRPO-EnCoT & 53.44 & 58.52 & 57.45 & 57.09 & 54.85 & 54.48 & 57.55 & 53.94 & 54.91 & 56.25 & 55.85 \\
        + GRPO-TransTest & 55.45 & \textbf{59.79} & \textbf{58.79} & \textbf{58.82} & \textbf{56.82} & \textbf{56.95} & \textbf{58.42} & 55.41 & \textbf{58.05} & \textbf{57.79} & \textbf{57.63} \\
        + TAPO-ChrF++ & \textbf{55.48} & 59.39 & 58.05 & 58.46 & 55.15 & 54.88 & 57.55 & \textbf{55.51} & 56.02 & 57.22 & 56.77 \\
    \midrule
        Llama & 24.30 & 37.77 & 32.22 & 31.68 & 24.80 & 21.56 & 30.58 & 24.10 & 24.23 & 29.45 & 28.07 \\
        + GRPO-EnCoT & 34.73 & 48.09 & 44.05 & 44.95 & 40.51 & 38.30 & 44.79 & 42.68 & 40.68 & 43.45 & 42.22 \\
        + GRPO-TransTest & 37.03 & \textbf{47.16} & 45.29 & 44.28 & 40.88 & 39.44 & 43.52 & 41.54 & 40.44 & 43.85 & 42.34 \\
        + TAPO-ChrF++ & \textbf{38.03} & 47.09 & \textbf{45.99} & \textbf{45.96} & \textbf{43.42} & \textbf{40.68} & \textbf{46.59} & \textbf{43.18} & \textbf{42.11} & \textbf{44.79} & \textbf{43.78} \\
    \bottomrule
    \end{tabular}
\caption{The detailed performance in each language on MMATH.}
\label{tab:mmath_results}
\end{table*}

\begin{table*}
    \centering
    \small
    \begin{tabular}{l|ccccccccccc}
    \toprule
        \textbf{Model} & \textbf{Bn} & \textbf{De} & \textbf{En} & \textbf{Es} & \textbf{Fr} & \textbf{Ja} & \textbf{Ru} & \textbf{Sw} & \textbf{Th} & \textbf{Zh} & \textbf{Avg} \\
    \midrule
        Qwen & 68.46 & \textbf{85.25} & \textbf{89.10} & \textbf{85.45} & \textbf{86.05} & 82.48 & \textbf{83.59} & 33.81 & 76.78 & \textbf{84.49} & 77.55 \\
        + GRPO-EnCoT & 69.36 & 81.49 & 85.61 & 81.18 & 81.98 & 80.13 & 79.78 & 56.73 & 75.54 & 80.99 & 77.28 \\
        + GRPO-TransTest & 68.78 & 80.23 & 84.06 & 79.24 & 80.70 & 78.11 & 79.49 & 59.03 & 74.43 & 79.86 & 76.39 \\
        + TAPO-ChrF++ & \textbf{72.38} & 83.96 & 86.75 & 83.10 & 84.31 & \textbf{82.51} & 82.51 & \textbf{65.96} & \textbf{77.88} & 82.71 & \textbf{80.21} \\
    \midrule
        Llama & 39.30 & 69.61 & 78.88 & 72.60 & 71.15 & 60.20 & 67.99 & 30.30 & 56.75 & 67.49 & 61.43 \\
        + GRPO-EnCoT & 62.85 & 77.11 & 83.91 & 77.55 & 77.81 & 73.26 & 75.68 & 67.16 & 67.91 & 74.21 & 73.75 \\
        + GRPO-TransTest & 67.18 & 79.03 & 83.84 & 78.45 & 78.95 & 75.29 & 77.84 & 71.90 & 71.04 & 76.70 & 76.02 \\
        + TAPO-ChrF++ & \textbf{68.98} & \textbf{79.50} & \textbf{84.73} & \textbf{79.36} & \textbf{79.81} & \textbf{77.09} & \textbf{77.68} & \textbf{73.26} & \textbf{73.28} & \textbf{78.84} & \textbf{77.25} \\
    \bottomrule
    \end{tabular}
\caption{The detailed performance in each language on MSVAMP.}
\label{tab:msvamp_results}
\end{table*}

\end{document}